\newif\ifisfinal\isfinaltrue
\documentclass[10pt,twocolumn]{article}

\usepackage[init=true]{sdoh}
\usepackage{zlhref}
\usepackage{zlsec}
\usepackage{zlfig}
\usepackage{zllist}
\usepackage{zltable}
\usepackage[
  bibname=sdoh,
  biboptions={natbib=true,backend=biber,hyperref=true,style=nature},
  superscript=true
]{zvgenmed}

\usepackage[margin=2.5cm]{geometry}

\zaveninit%

\zeauthors%

\usepackage[T1]{fontenc}
\usepackage[scaled=.95]{helvet}

\zttabvarsize[\tabsize]{.85}

\usepackage{widetext}
\renewenvironment{abstract}{%
  \begin{widetext}
    \vspace{-1.6cm}
    \centering
    \begin{minipage}{0.85\textwidth}
      \centerline{\textbf{\abstractname}} \small}{%
    \end{minipage}
  \end{widetext}}

\usepackage{dblfloatfix}

\begin{document}
\zavenbegindoc%
\begin{abstract}

  \sdoh\ are economic, social and personal circumstances that affect or influence an individual's health status.  \sdohs\ have shown to be correlated to wellness outcomes, and therefore, are useful to physicians in diagnosing diseases and in decision-making.  In this work, we automatically extract \sdohs\ from clinical text using traditional deep learning and \llms\ to find the advantages and disadvantages of each on an existing publicly available dataset.
  Our models outperform a previous reference point on a multilabel \sdoh\ classification by \diffGue\ points, and we present a method and model to drastically speed up classification (\twoStepSpeedUp{}X execution time) by eliminating expensive \llm\ processing.  The method we present combines a more nimble and efficient solution that leverages the power of the \llm\ for precision and traditional deep learning methods for efficiency.  We also show highly performant results on a dataset supplemented with synthetic data and several traditional deep learning models that outperform \llms.  Our models and methods offer the next iteration of automatic prediction of \sdohs\ that impact at-risk patients.

\end{abstract}

\acresetall % reset the "memory" of acro

\zssec[intro]{Introduction}

\sdoh\ greatly affect health, well-being and quality of life.  Safe housing, job opportunities, discrimination, and environmental factors are just a few examples \sdoh\ types.  Adverse \sdohs\ have immediate and profound affects on patient's health, such as air quality's negative affect on one's respiratory system or access to healthcare for those without income stability.  The impact of \sdohs\ on health outcomes have been shown to be helpful in prediction of diabetes mellitus~\cite{hill-briggsSocialDeterminantsHealth2021}, recurring diabetic keto-acidosis\cite{lyerlaRecurrentDKAResults2021}, and prolonged hospital stays~\cite{keenanSocialFactorsAssociated2002}.

Although \ehr\ systems have incorporated \sdohs\ as structured data, many systems have not, or represent the data with highly varying formats~\cite{liRealizingPotentialSocial2024,wangDocumentationReviewSocial2021,goldAdoptionSocialDeterminants2018}.  To address these challenges, many have turned \nlp\ methods to automatically extract \sdohs\ from clinical text.  The current \ac{sota} for this type of extraction incorporates \llms\ in either a few-shot learning or supervised-fine tuning setting~\cite{guevaraLargeLanguageModels2024,lituievAutomaticExtractionSocial2023}.

\summaryResultTab[t]{\tabsize} % table: Multilabel model performance by average

\byLabelResultTab[b]{\tabsize} % table: Multilabel model performance by label

We build on the work of \getal\ by training a variety of similar models on their publicly released datasets\footnote{We can not compare directly as \getal\ did not release their model, training dataset or prompts. In their work, the \mimicname\ and synthetic datasets were used for testing only.}  \ztsee{datasetStatsTab}.  The first dataset is subset of the \mimic\ corpus\cite{\mimicct}, \mimicdef.  This dataset consists of 5,355 sentences taken from \mimic\ clinical notes that were annotated for zero or more \sdohs.  A synthetic dataset generated from \llms\ was also released, containing one or more \sdohs, with 588 sentences annotated with at least one \sdoh.  The synthetic dataset was intended to provide a larger training set to boost results on the test dataset (\mimic).

Our experiments test Llama models~\cite{\llamact} for multilabel classification, and encoder-only \bert~\cite{\bertct} models for both multilabel and binary classification.  We experiment with several feature combinations in our traditional models and our \llm\ models with few-shot and supervised fine-tuning settings.  After error and performance analysis, we integrated both \llms\ and traditional deep learning \nlp\ models to exploit the best of both worlds.

Our binary traditional deep learning models performed inference up to \twoBinaryToLlamaSpeedup\ times faster than our slowest \llm\ \ztsee{resultLatencyTab}.  Based on this observation, we adopted a new \twostep\ model that predicts sentences with at least one \sdoh.  For those that do, we use the predictive power of Llama for multilabel classification.  The models were evaluated with a 10-fold cross-validation and with train, test and validation splits.

As small syntax changes can increase the variability of the generated text~\cite{wangSelfConsistencyImprovesChain2022}, we release our prompts for reproducibility (see \zsapxref{few-shot-prompt} and \zsapxref{train-prompt}).  We also release our data splits with ordering and our source code \zsseesec{data-avail}.  Next, we report on the results of all the classifiers, the performant \twostep\ classifier and the surprisingly positive effects of the synthetic dataset.
 % includes abstract
\zssec{Results}

We compare our results to \getal, but only as a reference point and not as test-only datasets.  We show that the our \twostep\ classifier is \diffGueTwoStep\ macro F1 points more performant than the reference point baseline \ztsee{summaryResultTab} and \twoStepSpeedUp{} times faster than the smallest \llm\ \ztsee{resultLatencyTab}.  Furthermore, our models show improvement with the synthetic data added to the \mimic\ data (we call this the \mimthetic\ dataset) compared to previous work, which shows a decrease in performance~\cite{\sdohidct}.

\zssubsec[model-perf]{Multilabel classifier}

\ztRef{summaryResultTab} gives the performance metric averages and \ztref{byLabelResultTab} gives performance metrics by labels on the \mimic\ and \mimthetic\ datasets.  The highest performing model by macro F1 (\bestAveMimicMacro) is the \bestAveMimicModel\ model.  The traditional deep learning encoder-only \textit{RoBERTa Base} model shows the best weighted average F1 on the \mimic-only dataset but falls behind all fine-tuned \llms.  The traditional model trails \diffLamSmall\ points behind the \llamaeight\ model and \diffGueTrad\ points behind the \getal\ baseline reference (\gueMFScore).

\ztRef{multiCrossFoldMimicStatsTab} shows the results of the 10-fold cross-validation of the traditional deep learning multilabel classifier on the \mimthetic\ dataset.  The weighted F1 score (\mlMimtheticCvMeanWeightedFScore) is stable, but the macro F1 score (\mlMimtheticCvMeanMacroFScore) is relatively low comparatively.  This statistic is illuminating as the classifier performs better by random variation on the test split (\bestAveMimtheticMacro\ in \ztref{summaryResultTab}).

\multiCrossFoldMimicStatsTab[t]{\tabsize}

\zssubsec[binary]{Binary classifier}

The label results from the best performing binary classifier taken from the ablation study is given in \ztref{binByLabelResultsTab}.  It has a lower performance with the minority label (sentences having at least one \sdoh).  However, a 10-fold cross validation with \cvrepeats\ repeats yields an average macro F1 of \binaryMimicCvMeanMacroFScore\ \ztsee{binCrossFoldMimicStatsTab}.  The results show a high confidence of this statistic with a standard deviation just over a half a point across all 50 tests.

\binByLabelResultsTab{\tabsize} % table: Multilabel model performance by label

\binCrossFoldMimicStatsTab[t]{\tabsize}

\zssubsec[tsres]{\Twostep\ classifier}

The \twostep\ classifier performs within a weighted F1 point of the best model (traditional deep learning \roberta) and within 5 points of the best macro F1 score (\llamaeight) as shown in \ztref{summaryResultTab}.  The classifier uses the traditional deep learning model for the negative label (\nosdoh), and the recall is significantly lower macro recall compared to precision, shows it struggles with false negatives at the binary level.  This is also reflected in the comparatively low macro recall from the cross-validated results in \ztref{multiCrossFoldMimicStatsTab}.

\zfaddtc[t][]{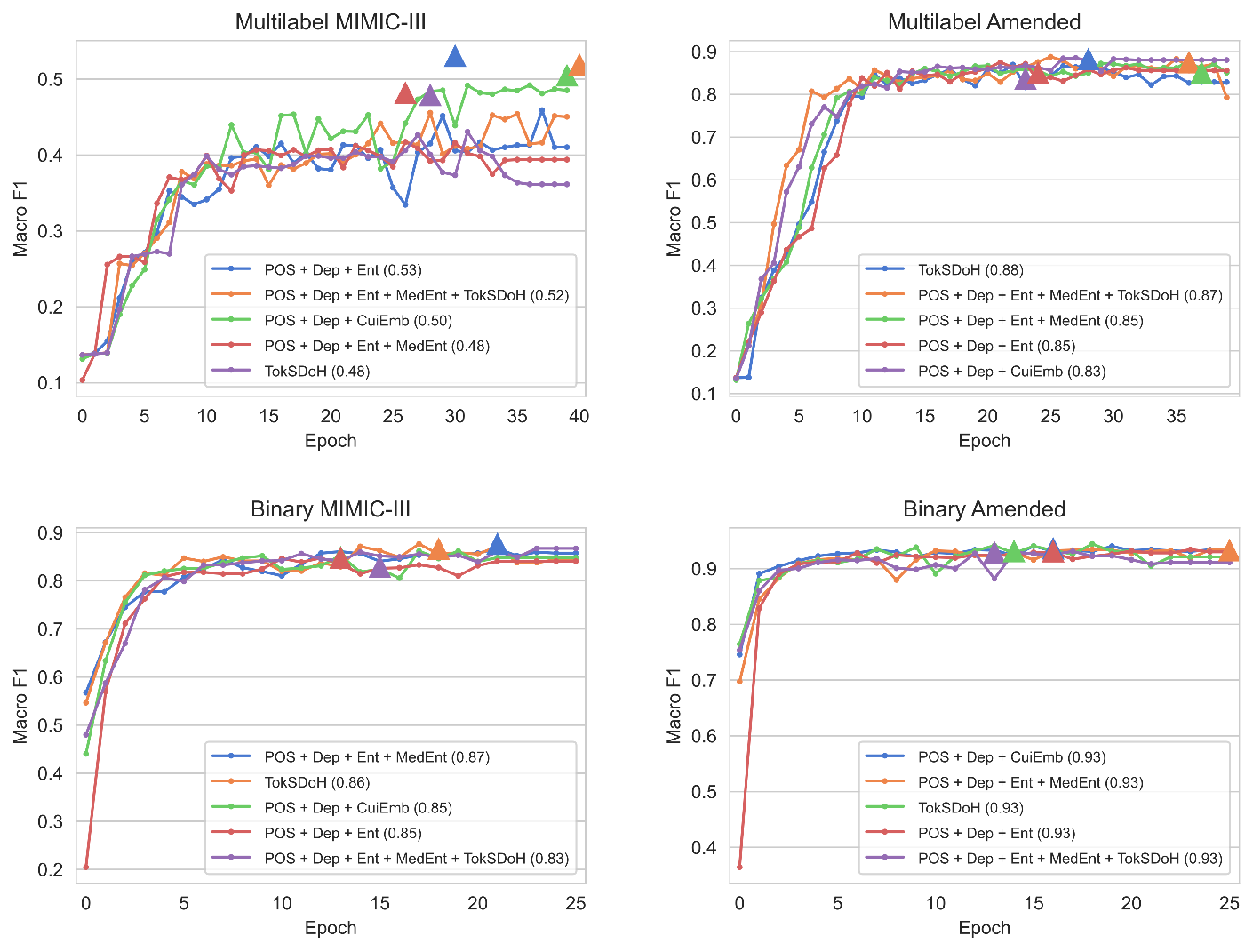}[Traditional model ablations]{The multilabel and binary model feature ablations by macro F1 performance are shown for the \mimic\ and \mimthetic\ validation sets.  The triangles represent the best performing macro F1 on the training set (also listed in the legend) and converged epochs on the X-axis.  Features are \scol{P}{art}-\scol{O}{of}-\scol{S}{each} tag, \scol{Dep}{endency} head tree depth, named \scol{Ent}{ity}, \scol{Med}{ical} named \scol{Ent}{ity}, \scol{Cui Emb}{bedding} and \scol{Tok}{en-level}\scol{SDoH}{}.}

\zssubsec[err]{Error analysis}

Our traditional deep learning multilabel classifiers failed for labels \texttt{Housing} and \texttt{Transportation} on the \mimic\ dataset.  These labels have only three occurrences each \ztsee{mimicCorpusTab}.  The \twostep\ classifier gets the single \texttt{Housing} instance correct, which means the binary traditional classifier learned that it was \sdoh\ positive and the \llm\ correctly classified it but still failed on the \texttt{Transportation} label.  However, all of our \llms\ achieved a non-zero score on all labels and outperformed the \getal\ reference models for most labels.  Furthermore, our models perform better on every label on the \mimthetic\ dataset.

The method of parsing of the \llm\ output might negatively affect performance.  A complex regular expression was needed to parse the noisy \llm\ output.  Models hallucinated in the few-shot setting (output on the \mimic\ dataset was more noisy than on the \mimthetic), but were more consistent on the fine-tuned models.  Of course, this was not an issue for the traditional deep learning model as its output layer directly predicted each label.  However, coupling traditional models with \llms\ can have consequences, such as with the \twostep\ classifier.

The \twostep\ classifier uses the binary classifier to predict if a \sdoh\ is present in a sentence.  When it predicts the presence of one or more \sdohs, it uses the \llamaeight\ model to assign labels.  As noted in \zssecref{tsres}, the binary classifier has a lower recall than precision, but the \llamaeight\ model's recall is significantly higher \ztref{summaryResultTab}.  This could be propagation error from false negative \sdoh\ classifications.  However, \ztref{multiCrossFoldMimicStatsTab} shows a very similar macro precision and recall, so the \llm\ might not assign any labels since the classifier was trained with \nosdoh\ as a label.  The \twostep\ classifier may perform better using a \llm\ trained without negative \sdoh\ labels.

\zssubsec[ablation]{Ablation studies}

Our ablation studies include feature combinations on the traditional deep learning binary classifier.  Each feature combination is a model that learns jointly with the \roberta\ embeddings \zsseesec{nn}.  \zfRef{binary-ablation} shows the ablation of the features as macro average F1 performance over epochs of training the models on the validation set.  The performance for the feature combination on the test set are displayed with triangles and in parentheses in the legend.

We see very high variance of the multilabel feature combinations' test set scores across model type.  Part-of-speech tags, head dependency features, and named entities were the most useful on the \mimic\ dataset but the \letal\ token-level \sdoh\ feature was the most helpful for the \mimthetic\ dataset.

The binary models on the \mimthetic\ dataset are much less sensitive to the choice of feature set implying the \roberta\ embeddings are leveraged to take advantage of the added synthetic data.  Adding features appears to negatively affect models on the \mimic\ dataset for some combinations.  The binary model illustrates how adding features may lead to worse performance.  The feature combination that includes all features performs five points lower on the \mimic\ dataset compared to the \mimthetic\ dataset.

\zssec[discuss]{Discussion}

As shown in \ztref{summaryResultTab}, our fine-tuned \llamaeight\ classifier shows a significant improvement over the \getal\ reference baseline on the \mimic\ dataset.  Our \llm\ performance metrics were based on a 20\% held-out test set across the \mimic\ and \mimthetic\ datasets whereas \getal\ use the annotated \mimic\ data as a test set.

The \llm\ models perform better on the \mimthetic\ dataset over the \mimic\ dataset as seen in \ztref{summaryResultTab} and \ztref{binCrossFoldMimicStatsTab}, which demonstrates that the models learn from the synthetic data.  The \mimic\ dataset's leader model was the \llamaeight.  However, the traditional deep learning models were not far behind on the \mimthetic\ dataset suggesting that the \llms\ models better adapt to the smaller dataset.

The traditional deep learning multilabel model was trained using the same features as the binary model described in the \zssecref{ablation}.  As explained in \zssecref{model-perf}, the multilabel classifier performs well overall, but very poorly on the minority labels.  The classifier under-performed on our \llms\ by 14 macro F1 points, but outperformed all \llms\ in the weighted F1.

As mentioned in \zssecref{err}, it is clear the traditional multilabel model has difficulty with the label imbalance given it fails in predicting the \texttt{Housing} and \texttt{Transportation} labels as shown in \ztref{byLabelResultTab}.  This result pales in comparison to the fine-tuned \llms\ that have a non-zero performance with all labels.

Furthermore, we observe the traditional model's ability at detecting \sdohs\ with a higher macro recall (0.49) than precision (0.6).  However, the converse is true with all \llms\ on the \mimic\ dataset (this dataset includes sentences with no \sdoh).  These observations motivate a binary classifier that predicts whether a sentence has any \sdoh\ \zsseesec{discuss}.  The high weighted F1 and recall metrics motivated the \twostep\ model.

\zssubsec{Inference latency}

The models differ greatly in inference latencies, particularly between the traditional models and \llms.  Pipeline processing bottlenecks arise as the latency of a classifier grows with the large input size of input, such as with longitudinal notes.

Our binary classifier performs as well as the best \getal\ model on the \mimic\ dataset.  However, it inferences at a fraction of the time of the \llms\ as the model is a fraction of the size.  The traditional binary classifier is able to predict up to \binSentPerSec\ sentences per second on average compared to the fastest \llm\ that predicts \fastestLlmSentPerSec\ sentences per second \ztsee{resultLatencyTab}.  The \twostep\ classifier is not far behind with speed up of \twoBinaryToLlamaSpeedup{}X, which translates to a prediction rate of \twoStepSentPerSec\ sentences per second.

\resultLatencyTab[t]{\tabsize}

\zssubsec{\Twostep\ and binary classifiers}

The binary classifier achieved a F1 of \binaryMimicCvMeanMacroFScore\ on our test split of the \mimic\ dataset.  The stratified dataset contains 992 negative \sdoh\ labels and 73 positive labels.  This dataset imbalance explains the lower F1 score of 0.767 on the positive labels.  However, it is \twoBinaryToLlamaSpeedup{} times faster compared to the \llamaeight\ fine-tuned model.

Our experiments show that the binary classifier used as the first component in the \twostep\ classifier is relatively close in performance to the fine-tuned \llms.  The \twostep\ classifier yields a macro F1 of \twoStepMimicMacro, which is only two macro F1 points lower than the \llamasev\ model \ztsee{summaryResultTab}.  Considering the \twostep\ classifier is \twoStepSpeedUp{} times faster than the \llamaeight\ classifier, we believe it shows the best performance trade off for real-world clinical application.

\datasetStatsTab[b]{\tabsize}

\zssec[method]{Methods}

We elaborated on the models by \getal\ using their datasets.  We also added a new model (\twostep) that integrates both the traditional deep learning binary model for efficiency and a \llm\ for precision.

\zssubsec[data]{Data}

\mimicCorpusTab[t]{\tabsize}

\mimtheticCorpusTab[t]{\tabsize}

The \getal\ publicly available \mimic\ and synthetic datasets were used for all experiments \zsseesec{data-avail}.  We also combined these two datasets, which we call the \mimthetic\ dataset.  Each data point in the datasets is a sentence and the associated \sdoh\ labels.  The labels apply to sentences rather than tokens.

We split both the \mimic\ and \mimthetic\ datasets each using a multilabel iterative stratification~\cite{sechidisStratificationMultilabelData2011} \zhfooturl{library}{https://github.com/trent-b/iterative-stratification} across \sdoh\ classes.  \ztRef{datasetStatsTab} shows our splits on each dataset with splits by label for the \mimic\ dataset in \ztref{mimicCorpusTab} and the \mimthetic\ label splits in \ztref{mimtheticCorpusTab}.

\zssubsec[model]{Model development}

We trained two types of models: binary (determines if a \sdoh\ exists) and a multilabel that classified zero or more \sdohs.  These models included:
\begin{itemize}
\item Supervised-fine tuned \llm: classifies zero or more \sdoh\ labels
\item Traditional deep learning multilabel: classifies zero or more \sdoh\ labels
\item Traditional deep learning binary: predicts whether a sentence has at least one \sdoh
\item \Twostep\ that integrates the traditional binary classifier with a \llm
\end{itemize}

The traditional deep learning and LLM models were trained and developed on the same dataset.

\zssubsec[model:llm]{Large language models}

The \llamaeight\ and \llamasev~~\cite{\llamact} models were used for all \llm\ experiments.  We used the \getal\ \anonguide\ to engineer our prompts for two settings: few-shot (\zsapxref{few-shot-prompt}) and supervised-fine tuned (\zsapxref{train-prompt}).

The few-shot prompts included the entire definition of each \sdoh\ category from the annotation guide.  The few-shot prompts also included examples of each of the six categories with a simple explanation of the special label (\nosdoh) for missing \sdohs.  The supervised fine-tuned training prompts included one or two sentence synopsis from the annotation guide for each category.  All fine-tuned \llms\ were trained using \lora~(\acl{lora})~\cite{\loract} with a rank of 64, a learning rate of $5 \times 10^{-5}$, and dropout of 10\% for three epochs.

Our experiments included feature prompt injection~\cite{kanayamaIncorporatingSyntaxLexical2024} in both few-shot and supervised fine-tuned settings using the token-level \sdoh\ feature~\cite{lituievAutomaticExtractionSocial2023} and the \cui's preferred name \zsseesec{cui-feature}.  However, only the few-shot setting marginally improved, so the results are not reported.

\zssubsec[model:trad]{Traditional deep learning models}

All traditional deep learning models were trained for 40 epochs, but the model with the lowest validation loss was used for evaluation.  The multilabel classifier was trained with a learning rate of $1 \times 10^{-5}$ and the binary classifier with a learning rate of $6.5 \times 10^{-6}$.

\zssubsubsec[feature-eng]{Feature engineering}

The traditional deep learning models incorporated several combinations of features.  The \roberta~\cite{\robertact} base model transformer was used and enhanced to accommodate token-level features \zsseesec{nn}.  These included linguistic features were extracted from tokenized text using \spacyname~\cite{\spacyct} and biomedical entities extracted with \scispacyname~\cite{\scispacyct}.  The \zeciteshort{lituievAutomaticExtractionSocial2023} model was used to add the prediction as the \sdoh\ token feature.

These features were concatenated to the transformer's final layer output for fine-tuning.  One influential feature was an encoded medical concept extracted from the input sentences \zsseesec{cui-feature}.  A list of features and their descriptions are given in \ztref{featureDescTab}.

\featureDescTab[b]{\tabsize} % table: feature descriptions

\zssubsubsec[cui-feature]{Concept features}

% remind reader and redefine (if we haven't already)
\acreset{umls}\acreset{cui}

The \umls\ is \umlsdef~\cite{\umlsct}.  Each node in the \umls\ graph represents a \cui.  Each \cui\ has many properties, two of which are the \textit{preferred name} (a common name for the concept) and a \textit{definition} of the concept, which can be exploited to provide more context to the model for each token.

We considered two methods creating additional features using \cuis\ found in the clinical text.  The first was simply to use embeddings generated from the \cui's preferred name and definition.  We also considered adding \cuitovec~\cite{\cuitovecct} embeddings, which are 500D vectors trained from clinical text using the \wtv\ algorithm~\cite{\wtvct}.  However, \cuitovec\ feature sparsity was a concern as its vocabulary is a subset of \umls, which is then further conditioned by a less than perfect recall by the entity linker.

\zfadd[t][0.85\columnwidth]{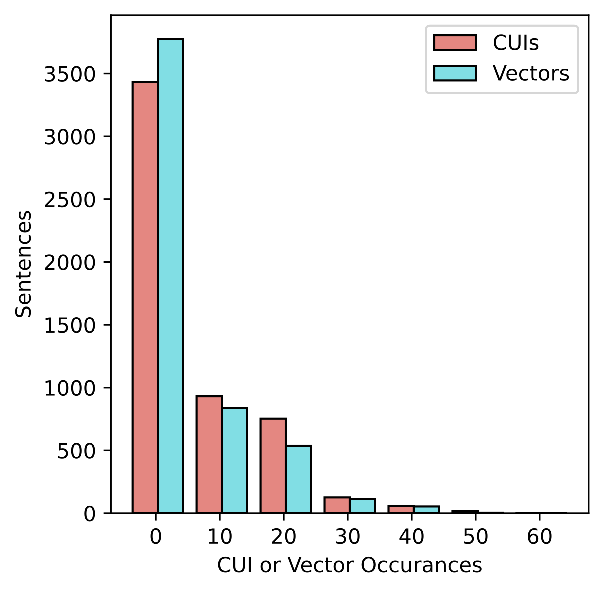}[\cui\ distribution]{Occurrences of \mimic\ dataset sentences with extracted concept identifiers and available mapped \cuitovec.  Each bar is the per sentence count (\zeie\ the 0-bar is the number of sentences with 0 \cuis\ extracted).}

\zfaddtc[][0.75\textwidth]{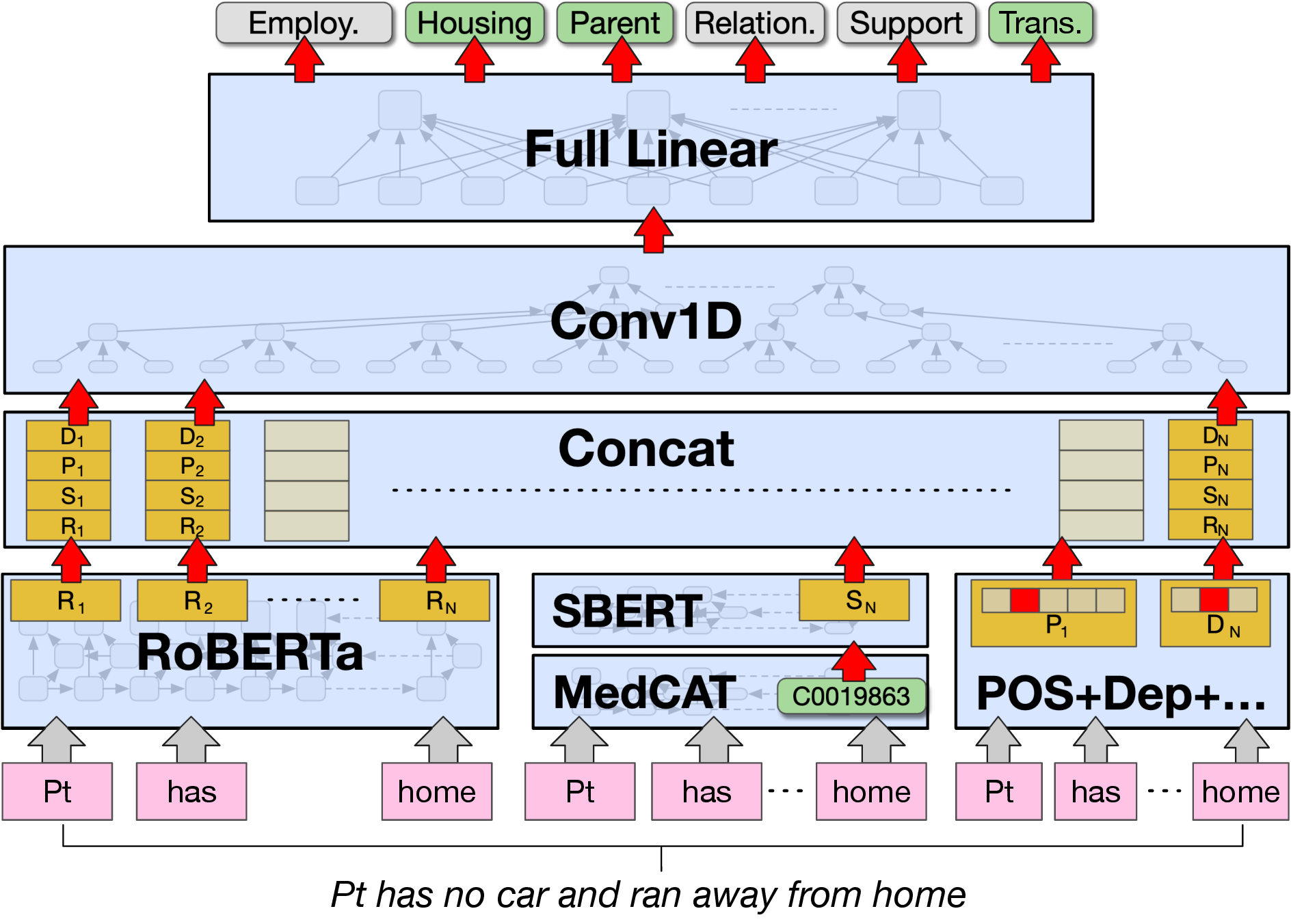}[Traditional multilabel model]{An example of a sentence processed through the traditional multilabel model.  The clinical text input is encoded by a \roberta, an entity linker and several linguistic taggers.  Each of these components' output is concatenated, passed through a 1D \ac{cnn} and decoded as zero or more labels.}

To make an informed decision, we computed a \cui\ and \cuitovec\ in-vocabulary distributions from the \mimic\ dataset.  \cuis\ were extracted using The \medcat~\cite{\medcatct} entity linker and were then assigned a numerical count per sentence.  As shown in the first bar in \zfref{cui-occurrences}, the number of sentences with no \cui\ is 3,434 (65\% of the dataset) and the \cuitovec\ embeddings available for those are even more scarce.  However, given that 1,887 sentences (35\% of the dataset) were found to have at least one \cui, we opted for a middle-ground solution to encode the \cui\ properties and leave the implementation of the \cuitovec\ features for future work.

To encode of the \cui\ properties, a \acl{sbert} (\sbert) model~\cite{\sbertct} was used to create embeddings.  We give the model a way to relate from medical concepts to \sdohs\ semantically as \sbert\ models embed text in Euclidean space.  Embedding the \cui\ with \roberta\ would add little or add redundancy in cases where the properties are close or identical to the text.

A clinically trained \sbert\ model~\cite{dekaImprovedMethodsAid2022} was used for the \cui{}s' embedding.  Only the static embeddings from a forward inference were used due to memory constraints of fine-tuning two (the \sbert\ and \roberta) models in parallel.  Text in the form \texttt{<preferred name>:<definition>} was used as input to the \sbert\ model and repeated (stacked) the embeddings across all \wpts~\cite{\wptct} (\wptdef) tagged by the entity linker for each concept.  Next, we explain how \cuis\ are vectorized and used with examples.

\zssubsubsec[nn]{Neural architecture}

The traditional deep learning multilabel and binary models use the same neural network architecture.  Only the output layer differs in the neuron cardinality: the multilabel model has one for each \sdoh\ label (6) and the binary model has one.

% example is present tense
\zfRef{trad-model} shows the three components that take the sentence as input:
\begin{itemize}
\item \textbf{\ac{roberta} (base) model's last layer}: The sentence is first tokenized and applied to the model.
\item \textbf{\cui\ extraction}: First \medcat\ links tokens to concepts.  The \cui\ for ``Homelessness'' (\texttt{C0019863}) is linked to the tokens ``ran away'' using preferred name ``Ran away, life event'' \zsseesec{cui-feature}.  This text is then used as input to \sbert, but the parameters of the \sbert\ model are not updated \zsseesec{cui-feature}.
\item \textbf{One-hot encoded features}: Vectors are encoded from enumerated linguistic values, such as part-of-speech tags. One-hot encoded features are: \texttt{POS}, \texttt{Dep}, \texttt{Ent}, \texttt{MedEnt} and \texttt{TokSDoH} \ztsee{featureDescTab}.
\end{itemize}

Each of these components take the sentence as input and are used in parallel to create enriched embeddings for each \wpt.  The components' output are then concatenated so that each \wpt\ has the \roberta\ last layer embedding, last layer \sbert\ embedding, and the one-hot encode vectors.  The concatenated tokens and features are then passed through a two layer 1D \ac{cnn}.  Finally, the last fully connected linear layer learns to decode the convolutional to \sdoh\ label.  A threshold for each neuron determines if the output is considered as present.

\zssubsubsec[twosteparch]{\Twostep\ model}

The \twostep\ model first uses the binary model to detect whether a sentence has one or more \sdohs.  For those that do, it then utilizes a larger more costly model with more precision for the multilabel classification\footnote{We joined the prediction data to measure performance and latency to simulate the classifier, but the implementation would be trivial.} such as a \llm.  Our results report the performance of the traditional deep learning binary model with the \llamaeight\ on the \mimic\ dataset.  We believe the results of the \twostep\ on the \mimthetic\ dataset would be higher, but leave this as a future work.

\zssec{Evaluation}

Instead of using the \getal\ \mimic\ and synthetic datasets for testing, we split them into train, validation, and test splits.  We then use these splits for all non-cross-validated tests to evaluate the traditional deep learning, \llamaeight\ and \llamasev\ models.  The evaluation was done on \mimic\ dataset and then again on the \mimic\ dataset with the synthetic data added in a similar fashion to the \getal\ experiments.  A 10-fold cross validation with \cvrepeats\ repeats was also used for evaluation.

The \scikitlearnurl\ multilabel and performance libraries were used to compute all metrics.  The model evaluation metrics included weighted, micro and macro average performance metrics on the multilabel iterative stratified~\cite{sechidisStratificationMultilabelData2011} splits.

All models were trained, tested, and validated on the splits described in \zssecref{data}.  The same train, validation, and test splits were used across the \llm\ and traditional models.  However, the traditional deep learning binary classifier on the \mimic\ dataset was evaluated with a 10-fold cross validation with \cvrepeats\ repeats\footnote{Cross-validation on the \llms\ were prohibitively expensive.}.  The \mimthetic\ multilabel classifier was also cross-validated with the same parameters set.

Micro, macro, and weighted averages were calculated using the following:
 %https://vitalflux.com/micro-average-macro-average-scoring-metrics-multi-class-classification-python/
\noindent{}Let:
\begin{zlpackeditemize}
\item $TP_i$, $FP_i$, $FN_i$ be the true positives, false positives, and false negatives for class $i$.
\item $N$ be the total number of classes.
\item $w_i$ be the weight for class $i$, typically the proportion of instances belonging to that class.
\end{zlpackeditemize}

\noindent{}Micro average:
\begin{eqnarray}
\textrm{P} &=& \frac{\sum_{i} TP_i}{\sum_{i} (TP_i + FP_i)} , \;
\textrm{R} = \frac{\sum_{i} TP_i}{\sum_{i} (TP_i + FN_i)} \nonumber \\
\textrm{F1} &=& \frac{2 \sum_{i} TP_i}{2 \sum_{i} TP_i + \sum_{i} FP_i + \sum_{i} FN_i} \nonumber
\end{eqnarray}

\noindent{}Macro average:
\begin{eqnarray}
\textrm{P} &=& \frac{1}{N} \sum_{i} \frac{TP_i}{TP_i + FP_i} , \;
\textrm{R} = \frac{1}{N} \sum_{i} \frac{TP_i}{TP_i + FN_i} \nonumber \\
\textrm{F1} &=& \frac{1}{n} \sum_{i=1}^{n} \left( 2 \times \frac{\frac{TP_i}{TP_i + FP_i} \times \frac{TP_i}{TP_i + FN_i}}{\frac{TP_i}{TP_i + FP_i} + \frac{TP_i}{TP_i + FN_i}} \right) \nonumber
\end{eqnarray}

\noindent{}Weighted average:
\begin{eqnarray}
\textrm{P} &=& \sum_{i} w_i \cdot \frac{TP_i}{TP_i + FP_i} , \;
\textrm{R} = \sum_{i} w_i \cdot \frac{TP_i}{TP_i + FN_i} \nonumber \\
\textrm{F1} &=& \sum_{i=1}^{n} w_i \left( 2 \times \frac{\frac{TP_i}{TP_i + FP_i} \times \frac{TP_i}{TP_i + FN_i}}{\frac{TP_i}{TP_i + FP_i} + \frac{TP_i}{TP_i + FN_i}} \right) \nonumber
\end{eqnarray}

\zssec[data-avail]{Data and code availability}

The \mimic\ and synthetic datasets are available on \getalsource[GitHub].  Our
\sdohllmmodeleight\ and \sdohllmmodelsev\ \llm\ models are hosted on the
HuggingFace Hub, and the traditional deep learning trained models are available
on \sdohtradmodel.
%\zssec[code-avail]{Code availability}
The source code for all experiments (including reusable Python libraries for
the research community) and our dataset splits are available on
\sdohsrcurl[GitHub].

\zssec[data-avail]{Acknowledgments}

This work was funded by an award from the Center for Health Equity using Machine
Learning and Artificial Intelligence (CHEMA) postdoctoral funding award at the
University of Illinois Chicago.

%\clearpage
\zavenenddoc%

\clearpage
\onecolumn
\zavenappendix%

\zsapx[few-shot-prompt]{Few-shot Prompt}
\begin{zlprompt}[\textwidth]{few-shot-prompt}{Few-shot prompt}{%
    Our prompt used for \sdoh\ prediction with definitions and examples take from
    the \zeciteauthor{\sdohidct} annotation guide.}
Classify sentences for social determinants of health (SDOH).  Definitions SDOHs are given in the below list:

* `housing`: The status of a patient's housing is a critical SDOH, known to affect the outcome of treatment. For the purposes of this annotation task, a sentence will be annotated as housing if it expresses a challenge relating to the place of residence of the patient. Please note that references to cities and towns, without mention of specific housing should NOT be considered an SDOH annotation. Attributes are Poor, Undomiciled, Other.

* `transportation`: This SDOH pertains to a patient's inability to get to/from their healthcare visits.. A patient being present at the treatment location, even if explicitly textually represented, or discussions of transportation unrelated to adequacy of transportation access, should NOT be considered an instance of Transportation SDOH. However, if there is a case of explicit textual representation that a patient is absent for treatment and that absence is due to transportation issues, then this IS considered an instance of Transportation SDOH. Attributes are Distance, Resource, Other.

* `relationship`: Whether or not a patient is in a partnered relationship is an abundant SDOH in the clinical notes. A sentence represents relationship status if it expresses evidence that a patient is married, in a partnership, divorced/separated, single, or widowed. Attributes are Married, Partnered, Divorced, Widowed, Single.

* `parent`: This SDOH should be used for descriptions of a patient being a parent to at least one child who is a minor (under the age of 18 years old). Tthe only evidence necessary for this SDOH is the existence of a patient's child under the age of 18. For the purposes of this task, "teenage children" can be considered minors. This SDOH category is binary and has no attributes, so the full annotation will just be the SDOH.

* `employment`: This SDOH pertains to expressions of a patient's employment status. A sentence should be annotated as an Employment Status SDOH if it expresses if the patient is employed (a paid job), unemployed, retired, or a current student. Attributes are Employed, Unemployed, Under-Employed, Disability, Retired, Student.

* `support`: This SDOH is a sentence describes a patient that is actively receiving care support, such as emotional, health, financial support.  This support comes from family and friends but not health care professionals.  The sentence must describe an act of care, participation in the patient's care, or an explicit statement that the person in the patient's life is "supportive", "caring for them", etc. In these cases, we wish to capture a patient's Social Support with this annotation. 

Here are some examples of "Sentence" input and "SDOH labels" you output:

### Sentence:Pt lives in Arlington.
### SDOH labels:```housing```

### Sentence:Pt lives 30mi away from hospital and and complains about needing to transfer three times each way.
### SDOH labels:```transportation```

### Sentence:Pt and her husband came into my office today.
### SDOH labels:```relationship```

### Sentence:Pt has 2 children ages 9 and 13.
### SDOH labels:```parent```

### Sentence:Pt works as an electrician in Rockland.
### SDOH labels:```employment```

### Sentence:Here today is Pt, her daughter, and supportive wife
### SDOH labels:```support```

Now classify the sentence with a comma-separated list of labels that are mostly likely to be present.  Only output the labels (or ```-``` for no SDOH found) surrounded by three back ticks.

### Sentence:{{ text }}
### SDOH labels:
\end{zlprompt}

\zsapx[train-prompt]{Training Prompt}
\begin{zlprompt}[\textwidth]{train-prompt}{Training prompt}{%
    Our prompt used for supervised fine-tuned training of \sdoh\ prediction
    with examples take from the \zeciteauthor{\sdohidct} annotation guide.}
Classify sentences for social determinants of health (SDOH).

Definitions SDOHs are given with labels in back ticks:

* `housing`: The status of a patient's housing is a critical SDOH, known to affect the outcome of treatment.

* `transportation`: This SDOH pertains to a patient's inability to get to/from their healthcare visits.

* `relationship`: Whether or not a patient is in a partnered relationship is an abundant SDOH in the clinical notes.

* `parent`: This SDOH should be used for descriptions of a patient being a parent to at least one child who is a minor (under the age of 18 years old).

* `employment`: This SDOH pertains to expressions of a patient's employment status. A sentence should be annotated as an Employment Status SDOH if it expresses if the patient is employed (a paid job), unemployed, retired, or a current student.

* `support`: This SDOH is a sentence describes a patient that is actively receiving care support, such as emotional, health, financial support.  This support comes from family and friends but not health care professionals.

* `-`: If no SDOH is found.

Now classify sentences for social determinants of health (SDOH) as a list labels in three back ticks. The sentence can be a member of multiple classes so output the labels that are mostly likely to be present.

### Sentence: {{ text }}
### SDOH labels: ```{{ labels }}```
\end{zlprompt}

\end{document}